\documentclass{article}
\usepackage{arxiv}

\usepackage[utf8]{inputenc} 
\usepackage[T1]{fontenc}    
\usepackage{hyperref}       
\usepackage{url}            
\usepackage{booktabs}       
\usepackage{amsfonts}       
\usepackage{nicefrac}       
\usepackage{microtype}      
\usepackage{lipsum}		
\usepackage{natbib}
\usepackage{doi}

\usepackage{amsmath}
\usepackage{amsfonts}
\usepackage{amssymb}
\usepackage{booktabs}       
\usepackage{array}          
\usepackage{subcaption}     
\usepackage{todonotes}      
\usepackage{color}

\newcolumntype{L}[1]{>{\raggedright\let\newline\\\arraybackslash\hspace{0pt}}m{#1}}
\newcolumntype{C}[1]{>{\centering\let\newline\\\arraybackslash\hspace{0pt}}m{#1}}
\newcolumntype{R}[1]{>{\raggedleft\let\newline\\\arraybackslash\hspace{0pt}}m{#1}}

\newcommand{\M}{\mathcal{M}}		
\newcommand{\MDP}[1][]{(\S_{#1},\A_{#1},\P_{#1},\Rwd_{#1},\gamma,\rho_0)}

\renewcommand{\P}{\mathsf{P}}		
\renewcommand{\S}{\mathcal{S}}		
\newcommand{\A}{\mathcal{A}}		
\newcommand{\Rwd}{\mathsf{R}}		

\DeclareMathOperator*{\argmax}{\textrm{argmax}}

\newcommand{\EE}[2][]{\mathbb{E}_{#1}\left[#2\right]}

\newcommand{\abs}[1]{\lvert#1\rvert}		
\newcommand{\card}[1]{\lvert#1\rvert}		

\newcommand{\R}{\mathbb{R}}

\newcommand{\eg}{\textit{e.g.},~}

\newcommand{\ie}{\textit{i.e.},~}

\newcommand{\thetitle}{IxDRL: A Novel Explainable Deep Reinforcement Learning Toolkit based on Analyses of Interestingness}
\renewcommand{\shorttitle}{IxDRL: A Novel Explainable Deep Reinforcement Learning Toolkit}
\newcommand{\thekeywords}{Explainable AI,  Reinforcement Learning, Interestingness Analysis, Global and Local Explanations,  Applications of xAI}
\newcommand{\theauthors}{Pedro Sequeira and Melinda Gervasio}

\newcommand{\theabstract}{
In recent years, advances in deep learning have resulted in a plethora of successes in the use of reinforcement learning (RL) to solve complex sequential decision tasks with high-dimensional inputs. However, existing systems lack the necessary mechanisms to provide humans with a holistic view of their competence, presenting an impediment to their adoption, particularly in critical applications where the decisions an agent makes can have significant consequences. Yet, existing RL-based systems are essentially competency-unaware in that they lack the necessary interpretation mechanisms to allow human operators to have an insightful, holistic view of their competency. Towards more explainable Deep RL (xDRL), we propose a new framework based on analyses of \emph{interestingness}. Our tool provides various measures of RL agent competence stemming from interestingness analysis and is applicable to a wide range of RL algorithms, natively supporting the popular RLLib toolkit. We showcase the use of our framework by applying the proposed pipeline in a set of scenarios of varying complexity. We empirically assess the capability of the approach in identifying agent behavior patterns and competency-controlling conditions, and the task elements mostly responsible for an agent's competence, based on global and local analyses of interestingness. Overall, we show that our framework can provide agent designers with insights about RL agent competence, both their capabilities and limitations, enabling more informed decisions about interventions, additional training, and other interactions in collaborative human-machine settings.
}

\title{\thetitle}

\author{
\theauthors\\
SRI International\\
333 Ravenswood Ave., Menlo Park, 94025 CA\\
\texttt{pedro.sequeira@sri.com, melinda.gervasio@sri.com}
}

\date{}

\renewcommand{\headeright}{}
\renewcommand{\undertitle}{}

\hypersetup{
pdftitle={\thetitle},
pdfsubject={\theabstract},
pdfauthor={\theauthors},
pdfkeywords={\thekeywords},
}

\begin{document}

\maketitle              

\begin{abstract}
\theabstract
\end{abstract}

\section{Introduction}%
\label{Sec:Intro}

\begin{figure*}[b]
    \centering
    \includegraphics[width=\textwidth]{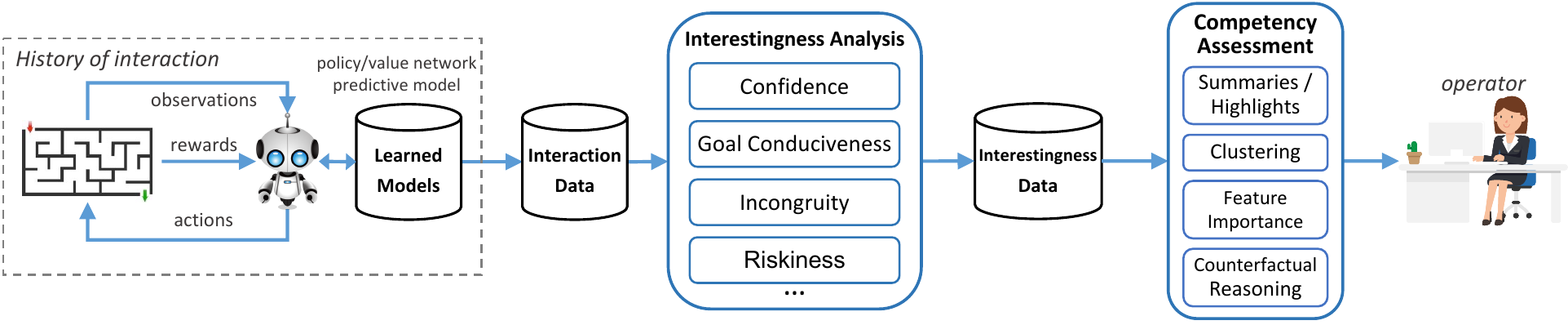}
    \caption{Our framework for analyzing the competence of deep RL agents through interestingness analysis.}%
	\label{Fig:Framework}
\end{figure*}

Reinforcement Learning (RL) is a machine learning technique for training autonomous agents to perform complex tasks through trial and error interactions with dynamic and uncertain environments. Recently, deep RL has achieved phenomenal successes, allowing agents to achieve---and even surpass---the performance level of human experts on various tasks, \eg \citep{silver2018rl,vinyals2019alphastar,berner2019dota2}. In addition to solving complex tasks in simulated environments, RL has also been applied in real-life, industrial settings \citep[see][for a recent survey]{naeem2020rlapps}. However, an impediment to the wider adoption of RL techniques for autonomous control, especially in critical settings, is that deep learning-based models are essentially black boxes, making it hard to assess their competency in a task, and identify and understand the conditions that affect their behavior.

In deciding whether to delegate a task to an autonomous agent, a human needs to know that the agent is capable of making the right decisions under the various conditions to adequately accomplish the task. The challenge with an RL agent is that after being trained and deployed, it will always select one action at each step, as informed by its policy---but the why (and why not) behind its decisions cannot be retrieved from the agent's model.
And while we can test RL agents prior to deployment and gather statistics about their performance on the task, or identify the actions they will select under certain circumstances, a more complete understanding of agents' competence---both in terms of its capabilities and limitations---remains essential to their acceptance by human collaborators.

Instead of capturing the agent's competence in terms of the specific decisions it makes under certain circumstances or its performance according to particular external metrics, we seek to characterize competence through self-assessment (introspection) over its history of interaction with the environment to capture distinct aspects that help explain its behavior. 
To do that, we analyze trained RL policies along various \emph{interestingness} dimensions following the work in \citep{sequeira2019interestingness}, \ie information that has the potential to be ``interesting'' in helping humans understand the competence of an RL agent.
The interestingness analyses capture aspects of agent competence such as whether an agent is confident in its action selections, or whether it recognizes risky or unfamiliar situations, among other things. Overall, the goal is to direct end-users of the RL system towards appropriate intervention by providing deeper insight into the agent's competence, \eg identifying sub-task competency (the situations in which the agent is more/less competent), or highlighting the situations requiring more training or direct guidance.

In this paper, we describe a novel framework for \textbf{I}nterestingness analysis for e\textbf{x}plainable \textbf{D}eep \textbf{RL} that we refer to as \textbf{IxDRL}, and use it to investigate the various forms by which we can use interestingness analyses to better understand the competency of different deep RL agents. The focus is on the quantitative and qualitative analyses over interestingness data, and on the insights about an agent's behavior that our framework can provide to potential end-users of our system. 
Fig.~\ref{Fig:Framework} depicts 
our IxDRL framework for competency-aware deep RL agents.
The input to our system is a trained deep RL policy that provides a set of \emph{learned models} that depends on the underlying RL algorithm, but typically corresponds to policy and value deep neural networks. 
Then, we test the agent by deploying it in the environment a number of times under different initial conditions, resulting in a set of traces. As the agent interacts with the environment, we probe the learned models and collect various information about the agent's behavior and internal state, \eg the value attributed to a state, or the probability each action has of being selected; we refer to all this information as the \emph{interaction data}.
We then perform interestingness analysis along several dimensions, resulting in a scalar value for each timestep of each trace---the \emph{interestingness data}. 
Finally, we perform competency assessment based on interestingness using various techniques.
%
Our contributions are as follows:
\begin{itemize}
    \item A new set of interestingness dimensions designed to cover different families of deep RL algorithms. Our implementation of the interestingness framework is compatible with popular RL toolkits.%
    \footnote{The IxDRL toolkit code is available at: \url{https://github.com/SRI-AIC/ixdrl}.}
    \item A method to analyze an RL agent's behavior in a task by clustering traces based solely on interestingness. The method enables the identification of distinct competency-controlling conditions that in different behavior patterns.
    \item A method to discover which task elements impact agent’s behavior the most and under which circumstances. We conduct feature importance analysis via SHapley Additive exPlanations (SHAP) \citep{lundberg2017shap} to perform global and local interpretation for competency assessment.
\end{itemize}
%
We present the results of a computational study where we trained RL agents for three different scenarios of varying complexity: Hopper, Atari Breakout, and a custom combat task running on the StarCraft II (SC2) platform. We used our interestingness framework to analyze the resulting agents' behavior and task competencies. 
We show that by going beyond trying to capture what an agent will do when and how well it will do it, our interestingness framework can provide human operators and partners a more complete picture of an RL agent's competence.
Furthermore, the trace clusters (obtained based on interestingness data) expose disparate challenges resulting in distinct agent behaviors, allowing the identification of different sub-tasks within the general task and distinct competency-controlling conditions. In addition, our feature importance analysis via SHAP provides insights about the ``sources'' of interestingness in the environment that most impact the agent’s competence, while also helping explain the contributions of each task element on particular situations having ``abnormal'' values of interestingness. Altogether, the higher-level competency modeling enabled by interestingness can provide human users with a more complete understanding of an agent's competencies, allowing them to make better decisions regarding the agent's use, providing insights on the agent’s limitations in the task, and suggesting directions for improving agent behavior.

\section{Related Work}%
\label{Sec:RelatedWork}

In recent years, many approaches to explainable RL (xRL) have been proposed for explaining various aspects of learned policies \citep{heuillet2021xrl,puiutta2020xrl}. These include: identifying the regions of the input that most affect an agent's decisions \citep{zahavy2016understanding,greydanus2018visualizing}, providing example trajectories \citep{huang2019enabling}, highlighting key decision moments to summarize an agent's behavior \citep{amir2019,huang2018establishing,lage2019summary}, extracting high-level descriptions of an agent's policy \citep{dereszynski2011learning,hostetler2012inferring,sequeira2022camly2,koul2019fsm}, and generating counterfactual explanations to help in understanding an agent's behavior \citep{madumal2020causal,yeh2022counterfactuals,olson2021counterfactual}. Here we focus on the approaches that attempt to provide a more complete view of RL agent competence in a form understandable to humans. 

Some approaches try to identify key moments of the agent's interaction that help explain its behavior in the task. For example, \citet{amir2019} propose some heuristics to select important states that are then used to summarize an agent's policy. A similar heuristic is presented in \citep{huang2018establishing}. \citet{lage2019summary} take a slightly different approach to behavior summarization, identifying important trajectories to be shown to an end-user. 
Although these approaches provide deeper insight into agent competence than the previously work, they still provide only a limited view of competence and can only be applied to specific families of \emph{tabular} RL methods.

Some approaches focus on explaining behavior in the form of queryable models that let users analyze agent behavior under different conditions. For example, \citet{hayes2017policyexplanation} find correlations between conditions and actions and lets users query the model through templated questions around identifying conditions for actions, predicting what an agent will do, and explaining expectation violations. Meanwhile, \citet{vanderwaa2018contrastive} use simulation to predict sequences of future actions to answer user questions about the consequences of actions/policy. 

Another body of work attempts to characterize agent behavior by extracting diverse structural models to represent agent strategies. For example, \citet{dereszynski2011learning,hostetler2012inferring} use probabilistic graphical modeling to learn finite-state models of strategy for the StarCraft domain, while \citet{sequeira2022camly2} infer strategies in the form of logical task specifications from agent traces using information-theoretic techniques to capture the conditions under which different behaviors occur. These approaches are focused solely on characterizing agent behavior and do not attempt to capture other aspects of the agent's decision-making. 

In earlier work on interestingness analysis \citep{sequeira2020ixrl}, we performed a user study that involved showing users short video clips highlighting different interestingness moments of agent interactions with the environment for RL agents with different capabilities and limitations on the task. The results revealed that some dimensions are better than others at conveying agent competence and that the diversity of aspects captured by the different interestingness dimensions helped users better understand an agent's task competencies. This paper builds upon our previous work \citep{sequeira2019interestingness,sequeira2020ixrl}, expanding it through a novel set of analyses that can extract interestingness from deep RL policies and not just tabular data. In addition, we go beyond summarizing behavior through the extraction of highlights, making use of interestingness for behavior clustering and local and global explanations of agent competence.

\section{Interestingness Analyses}%
\label{Sec:Interestingness}

In this section we detail our framework for analyzing RL agent competence through interestingness, starting by providing the necessary background on RL.

\subsection{Reinforcement Learning}%
\label{Subsec:RL}

We are interested in characterizing the competence of RL agents \citep[we refer to][for a more thorough description of the problem and main approaches]{sutton2018rl}. RL is a machine learning technique allowing autonomous agents to learn sequential decision tasks through trial-and-error interactions with dynamic, uncertain environments \citep{sutton2018rl}. RL problems can be framed under the \emph{Markov decision process} (MDP) formulation \citep{puterman1994mdp}, denoted as tuple $\M=\MDP$, where: $\S$ is the set of environment states; $\A$ is the set of agent actions; $\P(s'\mid s,a)$ is the probability of the agent visiting state $s'$ after executing action $a$ in state $s$; $\Rwd(s,a)\in\R$ is the reward function, that dictates the reward that the agent receives for performing $a$ in $s$; $\gamma\in[0,1]$ is a discount factor denoting the importance of future rewards; $\rho_0$ is the starting state distribution.

The goal of an RL algorithm is to learn a policy, denoted by $\pi(a|s)$, mapping from states to actions, that maximizes the expected return for the agent, \ie the discounted sum of rewards it receives during its lifespan. The optimization problem can be formulated by $\pi^*=\argmax_\pi{\EE{\sum_t{\gamma^t R_t}}}$, where $R_t$ is the reward received by the agent at discrete timestep $t$, and $\pi^*$ is termed the optimal policy. Often, RL algorithms use an auxiliary structure while learning a policy called the value function, corresponding to $V^\pi(s)=\EE{\sum_t{\gamma^t R_t})|S_0=s}$, that provides an estimate of the return the agent will receive by being in state $s$ and following policy $\pi$ thereafter. In deep RL, policies and other auxiliary structures are represented by neural networks whose parameters are adjusted during training to change the agent's behavior via some RL algorithm. As indicated in Fig.~\ref{Fig:Framework}, we refer to all such networks optimized via deep RL as the \emph{learned models}.

\subsection{Interaction Data}%
\label{Subsec:InteractionData}

Given a trained policy, the next step in our framework is the extraction of \emph{interaction data} given a trained RL agent. We collect these data by ``running'' the agent in the environment for a number of times,%
\footnote{Without loss of generality, here we deal with episodic tasks.}
collecting samples from the environment and probing the learned models at each step. The result is a set of \emph{traces} comprising the agent's history of interaction with the environment from which the agent's competence will be analyzed.

Distinct RL algorithms make use of different models to optimize the agent's policy during training. Our framework extracts data from four main families of RL algorithms:%
\begin{description}
    \item [Policy gradient] approaches \citep[\eg][]{sutton2018rl,schulman2017ppo,haarnoja2018sac} that provide a stochastic policy, $\pi(a|s)$, \ie a function mapping from observations to distributions over the agent's actions; 
    \item[Value-based] approaches \citep[\eg][]{mnih2015dqn,schaul2015replay} that compute a (state) value function, $V(s)$, that indicates how good it is for the agent to be in a situation, or an action-value function, $Q(s,a)$, asserting the value of executing some action given a state;
    \item[Model-based] approaches \citep[\eg][]{chua2018pets,janner2019mbpo} that learn a model of the environment's dynamics, $\P(s',r|s,a)$, \ie a function mapping from observations and actions to expected next observation and reward, that is used as a surrogate of the environment to collect samples and update the agent's policy;     
    \item[Distributional RL] approaches \citep[\eg][]{bellemare2017distributional} whose auxiliary structures, \eg the $Q$ function, output distributions over values instead of point predictions, allowing the capture of uncertainty around the estimates.
\end{description}

Interaction data then comprises everything we can extract at each timestep given the learned models provided by the RL agent. In addition to this internal agent data, we also collect external (observable) data, \ie the reward received by the agent, the selected action, and the agent's observation.

\subsection{Interestingness Dimensions}%
\label{Subsec:Dimensions}

As mentioned earlier, the goal of interestingness analysis is to characterize RL agents' competence along various dimensions, each capturing a distinct aspect of the agent's interaction with the environment. The dimensions of analyses are inspired by what humans---whether operators or teammates---might seek when trying to understand an agent’s competence in a task. Each dimension provides distinct \emph{targets of curiosity} whose values might trigger a human to investigate the agent's learned policy further \citep{hoffman2018xai}. As such, our system provides a means for the agent to perform competency self-assessment, where we use the data resulting from the interestingness analyses to identify cases that a human should be made aware of, and where user input might be needed.

Computationally, an analysis is given interaction data for each trace as input (see Fig.~\ref{Fig:Framework}) and, for each timestep, produces a scalar value in the $[-1,1]$ interval denoting the ``strength'' or ``amount'' of competence as measured by that interestingness dimension. Further, with the goal of enabling the analyses to be computed online, \ie \emph{while} the agent is performing the task, we restrict analyses to have access only to the data provided up to a given timestep. 
%
In this work, we designed and implemented a novel set of interestingness analyses that are applicable to a wide range of tasks and cover most of the existing deep RL algorithms. Notwithstanding, each RL algorithm allows the collection of only a subset of the interaction data mentioned in Sec.~\ref{Subsec:InteractionData}, which results in only a subset of the interestingness analyses being performed for any particular RL agent. 
Our IxDRL Python implementation is compatible with popular RL toolkits, including RLLib \citep{liang2018rllib}, an open-source library offering support for production-level, highly distributed RL workloads ($20+$ RL algorithms), which can foster wider adoption of our toolkit.
We now describe the goal behind each interestingness analysis and their mathematical realization in our framework.
\begin{description}
    \item[Value:] characterizes the long-term importance of a situation as ascribed by the agent’s value function. It can be used to identify situations where the agent is near its goals (maximal value) or far from them (low value). Given the agent’s value function associated with policy $\pi$, denoted by $V^{\pi}$, we compute \emph{Value} at discrete timestep $t$ using: $\mathcal{V}(t)=2\left(V_{[0,1]}^\pi(s_t)\right)-1$, where $V_{[0,1]}$ is the normalized value function obtained via min-max scaling across all timesteps of all traces.
    \item[Confidence:] measures the agent's confidence in its action selection, helping identify good opportunities for requesting guidance from a human user or sub-tasks where the agent may require more training. For discrete action spaces, where a stochastic RL policy is applicable, corresponding to a discrete probability distribution over the possible actions, we compute \emph{Confidence} at timestep $t$ with $\mathcal{C}(t)=1-2J\left(\pi(\cdot|s_t)\right)$, where $J(X)$ is Pielou’s evenness index \citep{pielou1966evenness}, corresponding to the normalized entropy of a discrete distribution $X$, which is given by $J(X)=-\frac{1}{\log{n}}\sum_i{P(x_i)\log{P(x_i)}}$. Our implementation also computes confidence for continuous action spaces, where policies typically parameterize a multivariate Gaussian distribution. When that is the case, we use the relative entropy-based dispersion coefficient proposed in \citep{kostal2010dispersion} to replace the $J$ evenness index above.
	\item[Goal Conduciveness:] assesses the desirability of a situation for the agent given the context of the decision at that point, \ie the preceding timesteps leading up to the current state. Intuitively, this computes how ``fast'' the agent is moving towards or away from the goal. Decreasing values can be particularly interesting, but we can also capture large differences in values, which potentially identify external, unexpected events that would violate operator's expectations and where further inspection may be required. We compute \emph{Goal Conduciveness} at timestep $t$ directly from the first derivative of the value function with respect to time at $t$, namely using: $\mathcal{G}(t)=\sin\left(\arctan\left(\rho \frac{d}{dt}V_{[0,1]}(s_t) \right)\right)$, where the sine of the angle generated by the slope (in an imaginary unit circle centered at $V_{[0,1]}(s_t)$) is used for normalization and $\rho$ is a scaling factor to make the slope more prominent. In our implementation, we use $\rho=100$, and resort to a finite difference numerical method to approximate $\frac{d}{dt}V(s_t)$ given the value function of the $3$ previous timesteps.%
	\footnote{This corresponds to using the \emph{backward} finite difference coefficient with accuracy $2$ \citep{fornberg1988findiff}. A higher-order accuracy could be used if we wish to capture how the value function is changing for the computation of Goal Conduciveness, by using information from timesteps further back in the trace.}
	\item[Incongruity:] captures internal inconsistencies with the expected value of a situation, which may indicate unexpected situations, \eg where the reward is stochastic or very different from the one experienced during training. In turn, a prediction violation identified through incongruity can be used to alert a human operator about possible deviations from the expected course of action. Formally, we capture \emph{Incongruity} via the temporal difference (TD) error \citep{sutton2018rl}, \ie $\mathcal{I}(t)=r_t+\gamma V^\pi(s_t)-V^\pi(s_{t-1})$.%
	\footnote{This quantity is also known as the \emph{one-step TD} or \emph{TD(0) target}.}
	We then normalize $\mathcal{I}(t)$ by dividing it with the reward range observed from the task. 
	\item[Riskiness:] quantifies the impact of the ``worst-case scenario'' at each step, highlighting situations where performing the ``right'' vs. the ``wrong'' action can dramatically impact the 
 outcome. This dimension is best computed for value-based RL algorithms by taking the difference between the value of the best action, $\max_{a\in\A}{Q(a|s_t)}$ and the worst, $\min_{a\in\A}{Q(a|s_t)}$. However, here we use a policy-gradient algorithm that updates the policy directly. As such, we compute \emph{Riskiness} using $\mathcal{R}(t)=2\left(\max_{a_1\in\A}{\pi(a_1|s_t)} - \max_{a_2\in\A}{\pi(a_2|s_t)} \right)-1, a_1 \neq a_2$. This will usually result in a value similar to that of Confidence, but may help identify situations where \emph{one} of the actions is particularly undesirable (low-probability) compared to all the others, which can be used by an operator to further specify the conditions in which an action should never be executed.
    \item[Stochasticity:] captures the environment's \emph{aleatoric} uncertainty. This is the statistical uncertainty representative of the inherent system stochasticity, \ie the unknowns that differ each time the same experiment is run \citep{chua2018pets,janner2019mbpo}. Here, we capture the uncertainty around what happens when we execute the same action in the same state on different occasions. This analysis requires an algorithm that models the uncertainty of the agent’s environment, \eg, an algorithm implementing distributional RL \citep{bellemare2017distributional}. For learned models parameterizing discrete distributions of the $Q$-function, we compute stochasticity using $\mathcal{S}(t)=\frac{1}{\card{\A}}\sum_{a\in\A}{1-4\abs{D\left(Q^\pi\left(\cdot|s_t\right)\right)-0.5}}$, where $D(P)=2\sum_{k=0}^K{\frac{d_k}{K-1}}$ is Leik's ordinal dispersion index \citep{leik1966dispersion}, with $d_k=\begin{cases}c_k, & c_k\leq 0.5\\1-c_k, & \text{otherwise}.\end{cases}$, and $c_k=\sum_{i=0}^k{P(x_i)}$. This can be used to identify inherently stochastic regions of the environment, where different agent behavior outcomes may occur.
    \footnote{Our framework also computes stochasticity from models parameterizing continuous distributions, using an appropriate coefficient of variation in place of Leik's $D$.}
    \item[Familiarity:] estimates the agent's \emph{epistemic} uncertainty, corresponding to the subjective uncertainty, \ie due to limited data or lack of experience with the environment. We follow approaches in the model-based RL literature \citep{chua2018pets,janner2019mbpo,shyam2019modelexplore,pathak2019modelexplore} where epistemic uncertainty is measured by estimating the level of disagreement between different predictive models, forming an ensemble, that are trained independently, usually by random sub-sampling of a common replay buffer. Formally, an ensemble comprises a set of $K$ bootstrapped forward models, where each model, denoted by $f_\theta^k \left(s_{t+1},r_{t+1}|s_t,a_t \right)$, estimates the next-step state, $s_{t+1}$, and reward, $r_{t+1}$, given the current state, $s_t$, and action $a_t$ taken by the agent. We then compute the agent’s familiarity using $\mathcal{F}(t)=1-\frac{2}{K^2}\sum_{i,j}^{K}{d(s_{t+1}^i,s_{t+1}^j)}$, where $s_{t+1}^k$ is the predicted next state vector from model $k$ in the ensemble, and $d(x,y)\in[0,1]$ is any suitable standardized distance function between two state vectors predicted by the models. In our work we use the cosine distance. This analysis computes the level of (dis)agreement between the observations predicted by the models in the ensemble, and can thus be used to identify less-explored, unfamiliar parts of the environment where the agent might be more uncertain about what to do. It also identifies good intervention opportunities as it indicates regions that need further exploration.
    \footnote{Our implementation also computes familiarity from an ensemble of predictive models parameterizing distributions instead of outputting point predictions, in which case we use divergence measures between prediction distributions to replace for $d$.}
\end{description}

\begin{figure}[!tb]
    \centering
    \includegraphics[width=0.66\textwidth]{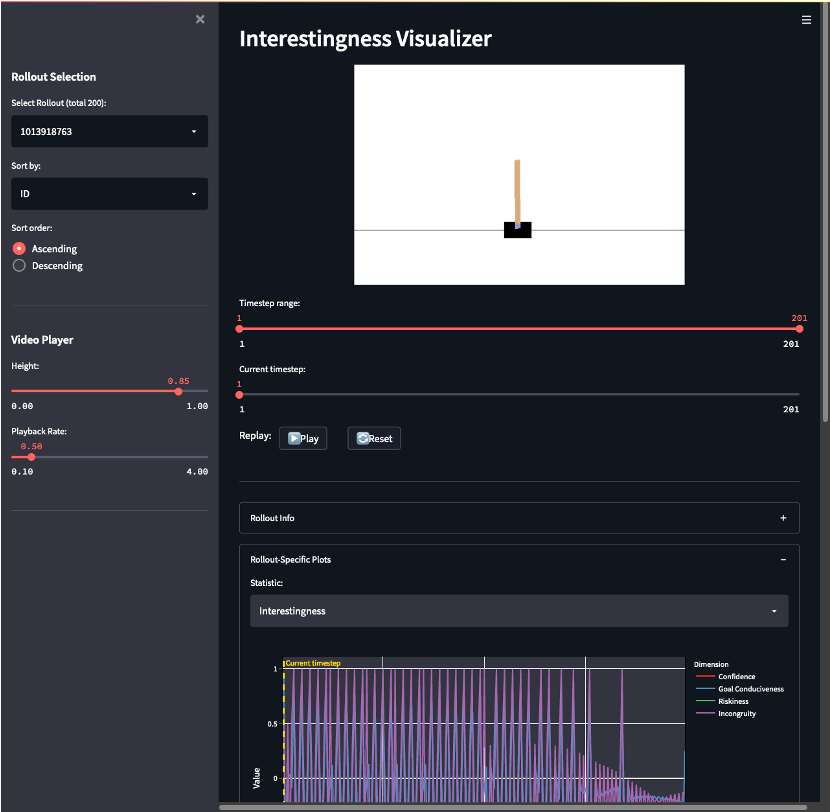}
    \caption{Interestingness and agent behavior visualization tool.}
	\label{Fig:IntUI}
\end{figure}

To help potential end-users of our framework analyze the behavior of agents and interestingness resulting from the several analyses, we built an interactive graphical tool (see Fig.~\ref{Fig:IntUI}) that allows visualizing videos of traces (replays), select a dimension and sort traces by mean interestingness value, and see the various plots automatically produced by our framework during interaction data extraction and interestingness analysis.

\section{Experiments and Results}%
\label{Sec:Experiments}

To validate our framework and understand the insights it can provide to potential end-users, we performed a computational study where we extracted interestingness data and applied different methods for interpreting the agent's competency in three distinct scenarios with varying degrees of control complexity. Although we do not perform a user study, our experiments are a necessary first step to identify the types of insights about RL agent competence that interestingness can provide. For each scenario, we describe the goal and environment dynamics, and the RL algorithm used to train each agent used for interestingness analysis.%
\footnote{All configurations used to train the RL agents, as well the data for each scenario, are available at: \url{https://github.com/SRI-AIC/23-xai-ixdrl-data}.}
We then illustrate the capability of our framework in analyzing the overall competence of the trained RL agents. Due to space restrictions, here we provide some statistics of the agents' behaviors and interestingness analyses and highlight only a few examples of the agents' capabilities and limitations.

\subsection{Environments}%
\label{Subsec:Envs}

\begin{figure*}[!tb]
    \centering
    \begin{subfigure}[b]{0.25\textwidth}
        \includegraphics[width=\textwidth]{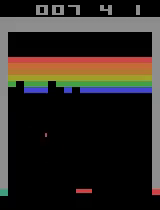}
        \caption{Breakout}%
        \label{Fig:Breakout}
    \end{subfigure}\hspace{5pt}%
    \begin{subfigure}[b]{0.27\textwidth}
        \includegraphics[width=\textwidth]{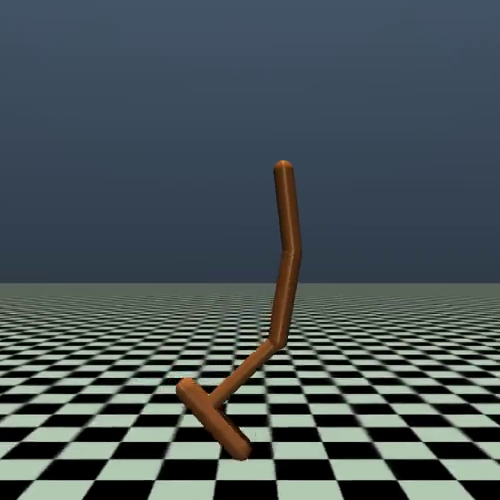}
        \caption{Hopper}%
        \label{Fig:Hopper}
    \end{subfigure}\hspace{5pt}%
    \begin{subfigure}[b]{0.45\textwidth}
        \includegraphics[width=\textwidth]{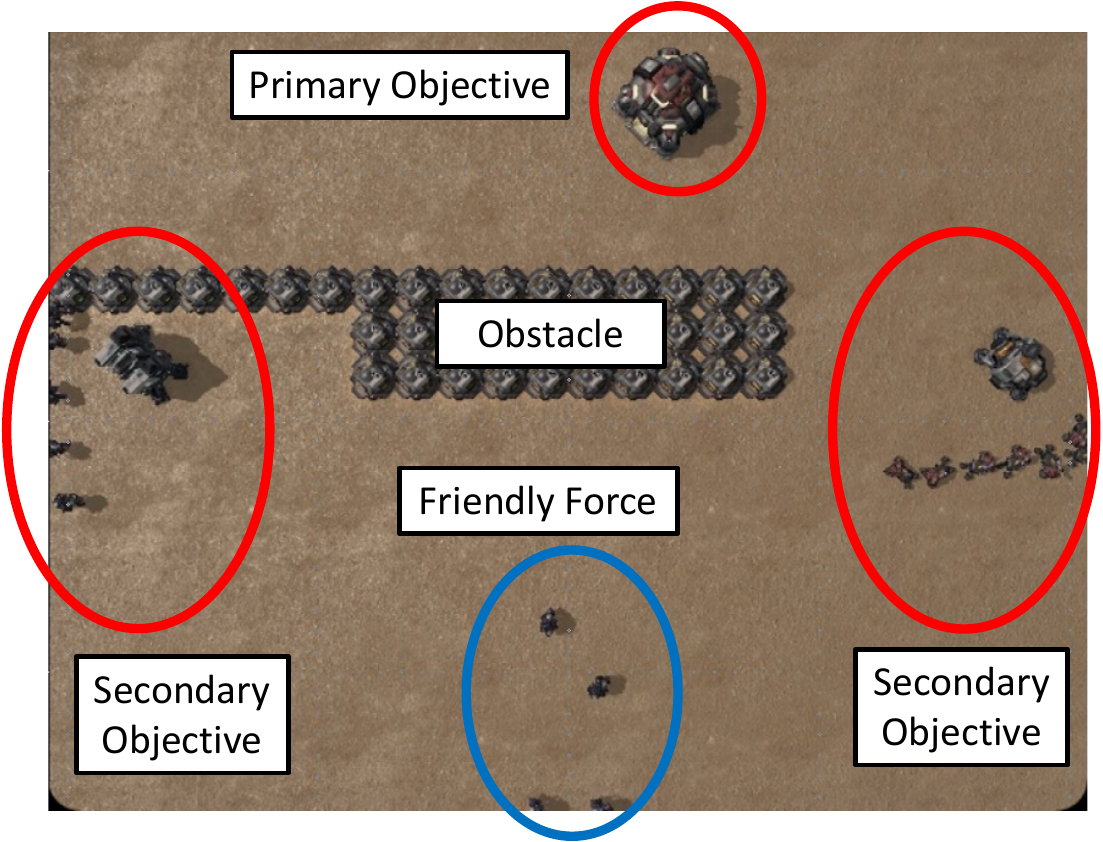}
        \caption{SC2 Combat}%
        \label{Fig:SC2}
    \end{subfigure}
    \caption{Screenshots of the different scenarios used in our experiments.}
	\label{Fig:Scenarios}
\end{figure*}

\subsubsection{Breakout}

The agent controls the horizontal movement of a paddle at the bottom of the screen (see Fig.~\ref{Fig:Breakout})%
\footnote{\url{https://gymnasium.farama.org/environments/atari/breakout/}}
 with the goal of hitting a bouncing ball into a colored brick wall at the top of the screen until it is fully destroyed. As more bricks are destroyed, the speed of the ball increases. The agent has five lives, and a life is lost whenever the ball falls into the ground (bottom part of the screen) without the paddle hitting it. Rewards (positive) are given when bricks are destroyed. We used the Arcade Learning Environment (ALE) implementation \citep{bellemare13arcade} with RLLib, where there are four discrete actions available (\emph{noop}, \emph{fire}, \emph{right}, \emph{left}), while the observations correspond to four time-consecutive stacked $84\times84$ grayscale frames transformed from the RGB game images. We trained an agent in RLLib for $2\times10^6$ timesteps using the distributional $Q$-learning approach as described in \citep{bellemare2017distributional} where the $Q$ function of each action is approximated using a discrete distribution over $51$ values in the $[-10,10]$ interval.

\subsubsection{Hopper}
The agent controls a one-legged simplified robotic structure (see Fig.~\ref{Fig:Hopper})%
\footnote{\url{https://gymnasium.farama.org/environments/mujoco/hopper/}}
that consists of four main body parts: the \emph{torso} at the top, the \emph{thigh} in the middle, the \emph{leg} in the bottom, and a \emph{foot} on which the entire body rests. The goal of the agent is to make hops that move the robot forward in the environment by applying torques (continuous values) on the three hinges connecting the four body parts, \ie there are three action factors. Observations are $11$-dimensional, consisting of positional values of the different body parts and their velocities. We trained an agent in the MuJoCo \citep{todorov2012mujoco} Hopper implementation using the Model-Based Policy Optimization (MBPO) algorithm \citep{janner2019mbpo}%
\footnote{We used the implementation at: \url{https://github.com/JannerM/mbpo}.}
that learns an ensemble of dynamics models, each predicting the parameters of a multivariate Gaussian distribution over the next-step observation and reward given the current state and performed action. The policy is optimized from rollouts produced by the learned models using Soft Actor-Critic (SAC) \citep{haarnoja2018sac}, where the idea behind using an ensemble is to prevent the policy from exploiting inaccuracies in any single model. The agent was trained for $15\times10^4$ timesteps, with $10^3$ initial steps of exploration used to train the dynamics models.

\subsubsection{StarCraft II Combat}%
To test our interestingness framework on a more complex control task, we implemented a custom combat scenario in \emph{Starcraft~II} \citep{blizzard2022starcraft} using the \texttt{pySC2} library \citep{vinyals2017starcraft} to interface with the game engine.%
\footnote{A more detailed description of our SC2 task is provided in \citep{sequeira2022camly2}.}
The agent controls the Blue force, which starts at the bottom of the map (see Fig.~\ref{Fig:SC2}). The agent's goal is to destroy the Primary Objective, which is a CommandCenter (CC) building located at the top of the map. The map is divided into three vertical ``lanes,'' each of which may be blocked by obstacles. The two side lanes contain Secondary Objectives, which are buildings guarded by Red forces. Destroying the Red force defending one of the secondary objectives causes the building to be removed from the map and replaced with additional Blue units (reinforcements), with the type of reinforcements determined by the type of the building destroyed. Different types of units have distinct capabilities. The starting Blue force consists of infantry (Marines or Marauders), but Blue can gain SiegeTanks (armored ground units) by capturing a Factory building, or Banshees (ground attack aircraft) by capturing a Starport. The Banshees are especially important because they can fly over ground obstacles and Red has no anti-air units to defend the primary objective. 

The initial state, including the number and type of Blue starting units, locations of obstacles, number of enemies, etc. is randomized to create a \emph{scenario} distribution. In addition, Red sometimes receives reinforcements at a random time step, and capturing a secondary objective sometimes does not grant Blue reinforcements. The agent observes the world as a top-down view corresponding to three semantic layers of size $192\times144$, each encoding a distinct property of objects/units at each location. The action space is factored over the four unit types that Blue can have. A joint action assigns an order separately to each unit type, and all units of that type execute the same order. The available orders are to \emph{target} the nearest Red unit of a specified type, to \emph{move} to one of $9$ fixed locations while ignoring any enemies encountered, or to \emph{do nothing}. The reward is a linear combination of five factors: a large reward for capturing the CC, rewards for gaining and losing Blue forces and for destroying Red forces in proportion to the resource cost of the units gained or lost, and a per-timestep cost. We trained an agent using a distributed implementation of the VTrace (actor-critic) algorithm \citep{espeholt2018impala} based on the open-source SEED-RL implementation \citep{espeholt2020seedrl}.

\subsection{Interaction Data and Interestingness Analysis}%
\label{Subsec:InterestingnessResults}

After training an RL agent for each scenario, we analyzed their competence using our IxDRL framework. We started by sampling $1{,}000$ traces by running each policy on the corresponding environment, which was randomly initialized where applicable. Overall, we configured the training regime such that the agents do not attain optimal performance in the corresponding task. This resulted in agents with different levels of competence. The Hopper agent attains a very good performance, resulting in a mean cumulative reward of $3{,}415\pm105$. Conversely, the Breakout agent's performance is not very consistent, attaining a mean cumulative reward of $110\pm143$. Often, it reaches a state where most bricks are destroyed but it is unable to destroy the remaining ones, bouncing the ball through empty space and walls in a continuous loop. As for the SC2 Combat agent, it achieved its primary objective (destroying the CC) $36\%$ of time and was defeated (no Blue units left) in $25\%$ of traces (remaining $39\%$ of traces timed out). The mean count of agent (Blue) units is $4.23\pm 3.47$ while the count for opponent (Red) units is $13.34\pm 5.59$. These results indicate that there appear to be task conditions for which the agents learned a strategy capable of achieving the task objectives, whereas others posed challenges that the agent could not overcome, and in which further training or guidance might be needed. These make the trained agents ideal candidates for interestingness analysis of competence.

\begin{table}[!ht]
    \centering
    \caption{The interestingness dimensions extracted for each scenario. The ``Total'' row indicates the total number of dimensions extracted. See text for details.}
    \begin{tabular}{l@{\hspace{10pt}} c@{\hspace{10pt}} c@{\hspace{10pt}} c@{\hspace{10pt}} }
    \toprule
        \textbf{Dimension} & \textbf{Breakout} & \textbf{Hopper} & \textbf{SC2 Combat} \\
    \midrule
        \emph{Value} & $\times$ & $\times$ & $\times$ \\ 
        \emph{Confidence} & $\times$ & $\times$ & $\times$ \\ 
        \emph{Goal Cond.} & $\times$ & $\times$ & $\times$ \\ 
        \emph{Riskiness} & $\times$ & $\times$ & $\times$ \\ 
        \emph{Incongruity} & $\times$ & $\times$ & $\times$ \\ 
        \emph{Stochasticity} & $\times$ & $\times$ & ~ \\ 
        \emph{Familiarity} & ~ & $\times$ \\
        \hline
        Total & $6$ & $10$ & $13$ \\
    \bottomrule
    \end{tabular}
    \label{Table:ScenarioDims}
\end{table}

After sampling the behavior traces, we collected interaction data as explained in Sec.~\ref{Subsec:InteractionData} for each of the $1{,}000$ sampled traces. Table~\ref{Table:ScenarioDims} shows which interestingness dimensions were produced for each scenario. Due to the types of observation and action spaces and RL algorithm used for each scenario, each scenario only allowed for the analysis of interestingness along particular dimensions. Only Hopper allows analysis for all the dimensions described in Sec.~\ref{Subsec:Dimensions} since, in addition to learning an ensemble of dynamics models, the models are distributional, which allows capturing both Familiarity and Stochasticity. For Hopper and the SC2 Combat scenario, because there are multiple action factors that the agent simultaneously controls at each timestep, Confidence can be computed for each factor separately. In the case of the SC2 task, Riskiness is also computed per action factor since the action space is discrete. We also compute the mean values for Confidence and Riskiness across action factors, thus resulting in a total number of dimensions higher than the number of analyses for these scenarios. 

\subsection{Overall Competence Assessment}%
\label{Subsec:CompetenceAss}

\begin{figure}[!tb]
    \centering
    \includegraphics[width=0.7\textwidth]{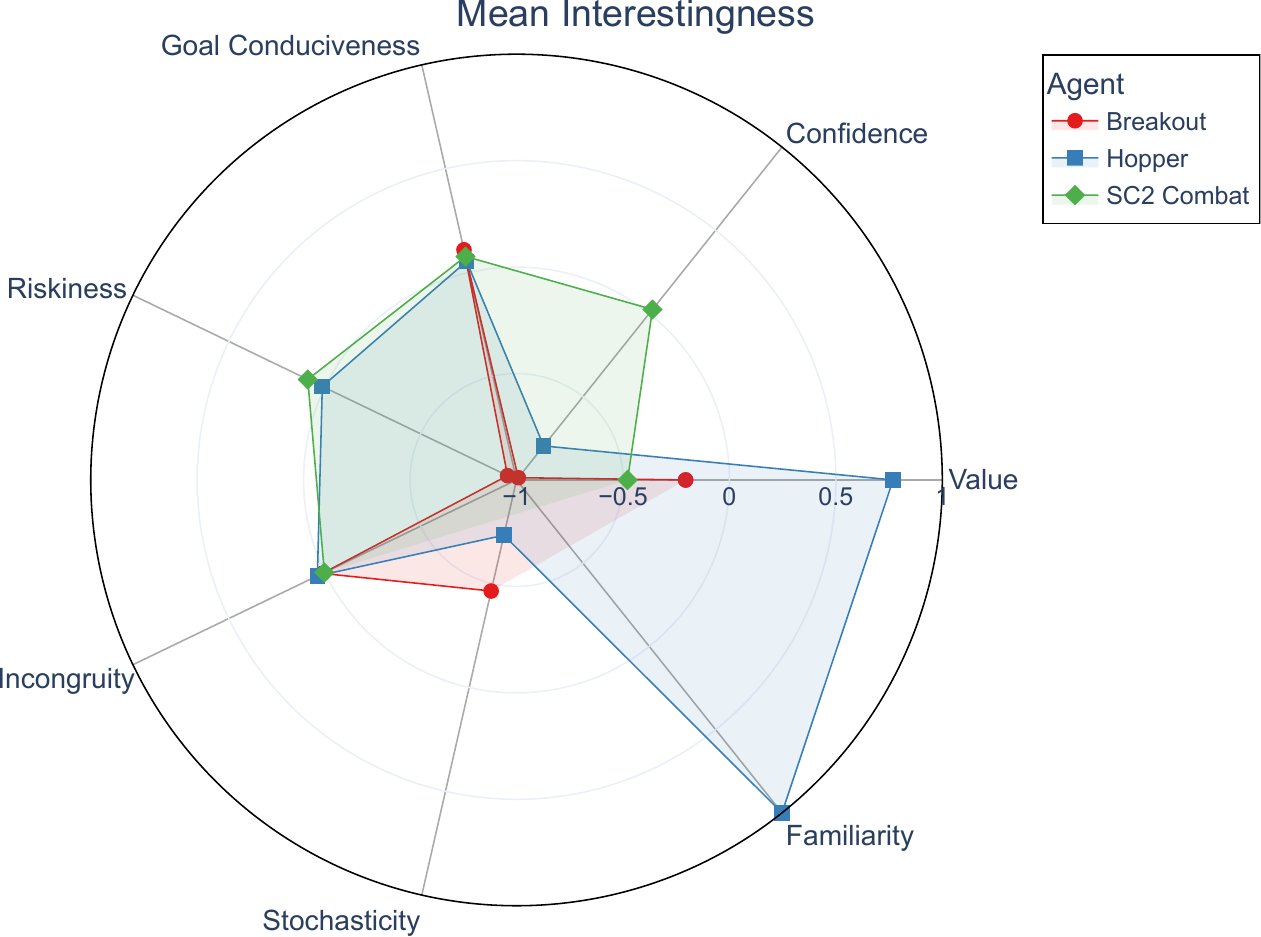}
    \caption{Interestingness profiles for each agent in the different scenarios.}
	\label{Fig:IntRadars}
\end{figure}

We start by assessing the agents' overall competency in the tasks as ascribed by the different interestingness dimensions. Fig.~\ref{Fig:IntRadars} shows a radar chart for the mean interestingness (across all timesteps of all traces) resulting from analyzing each agent. Each shape represents the interestingness profile or ``signature'' for the behavior of the corresponding agent. As we can see, different scenarios lead to distinct profiles. On average, the SC2 agent has a medium level of Confidence ($\mathcal{C}\approx0$) while the other agents are not so confident in the selection of actions. This is expected since the Breakout and Hopper tasks are more cyclic and the differences between the action values or associated probabilities are not that high (\eg moving the paddle left or right in Breakout only has impact when hitting the ball). In contrast, the SC2 task is more strategic and the agent needs to understand better which action to select at each step. Conversely, the Hopper agent attributes a higher Value to situations it encounters ($\mathcal{V}>0.5$) compared to the other agents who value situations negatively on average. This is also expected since the Hopper agent has the best performance overall, whereas the Breakout agent has a medium performance and the SC2 agent only values situations highly when in possession of the Banshees since, as explained earlier, they provide a winning advantage, 
although this does not happen frequently. Furthermore, all agents attribute on average a neutral level of Incongruity ($\mathcal{I}\approx0$), which is explained by the lack of reward stochasticity in all scenarios. The reward the agent receives is expected compared to the values associated with the previous and current observations. In the SC2 task, high Incongruity occurs when the agent receives the Banshees and/or the enemy receives reinforcements, but those situations seldom occur. For the same reason, the Goal Conduciveness results show that on average there are no abrupt changes to the agents' value functions ($\mathcal{G}\approx0$). As for Stochasticity, results show that the value attributed by the Breakout agent is very consistent after training (concentrated around a specific value) and that the probabilistic dynamics models learned via MBPO by the Hopper agent have a low variance. These results confirm that these scenarios are very deterministic ($\mathcal{S}\leq0.5$). Finally, on average, the Hopper agent attributes high Familiarity ($\mathcal{F}\approx1$) to the situations it encounters, showing that it is well experienced with its environment dynamics.

\begin{figure*}[!tb]
    \centering
    \begin{subfigure}[b]{0.50\textwidth}
        \includegraphics[width=\textwidth]{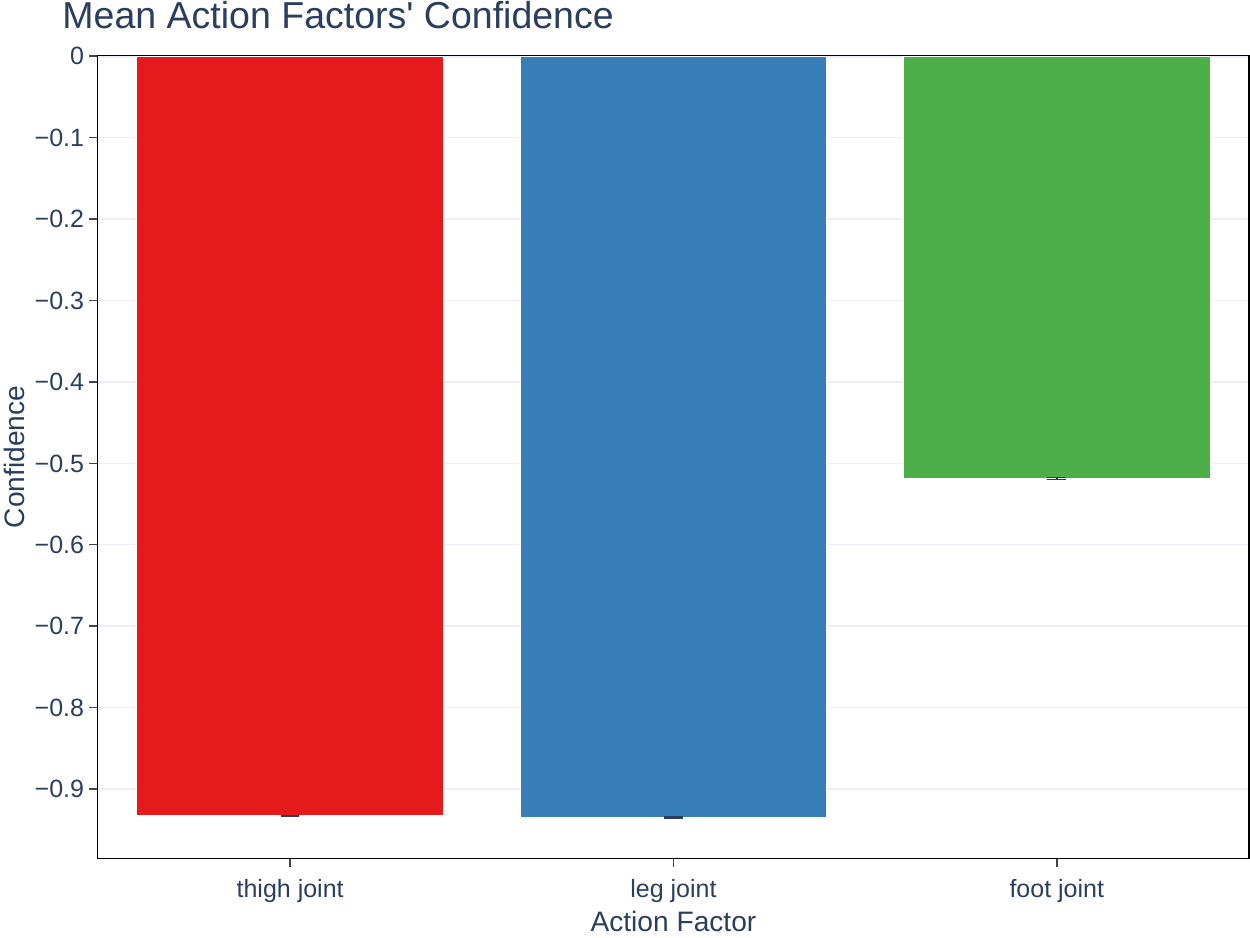}
        \caption{Action factors' Confidence.}%
        \label{Fig:HopperConfBar}
    \end{subfigure}\hspace{0pt}%
    \begin{subfigure}[b]{0.5\textwidth}
        \includegraphics[width=\textwidth]{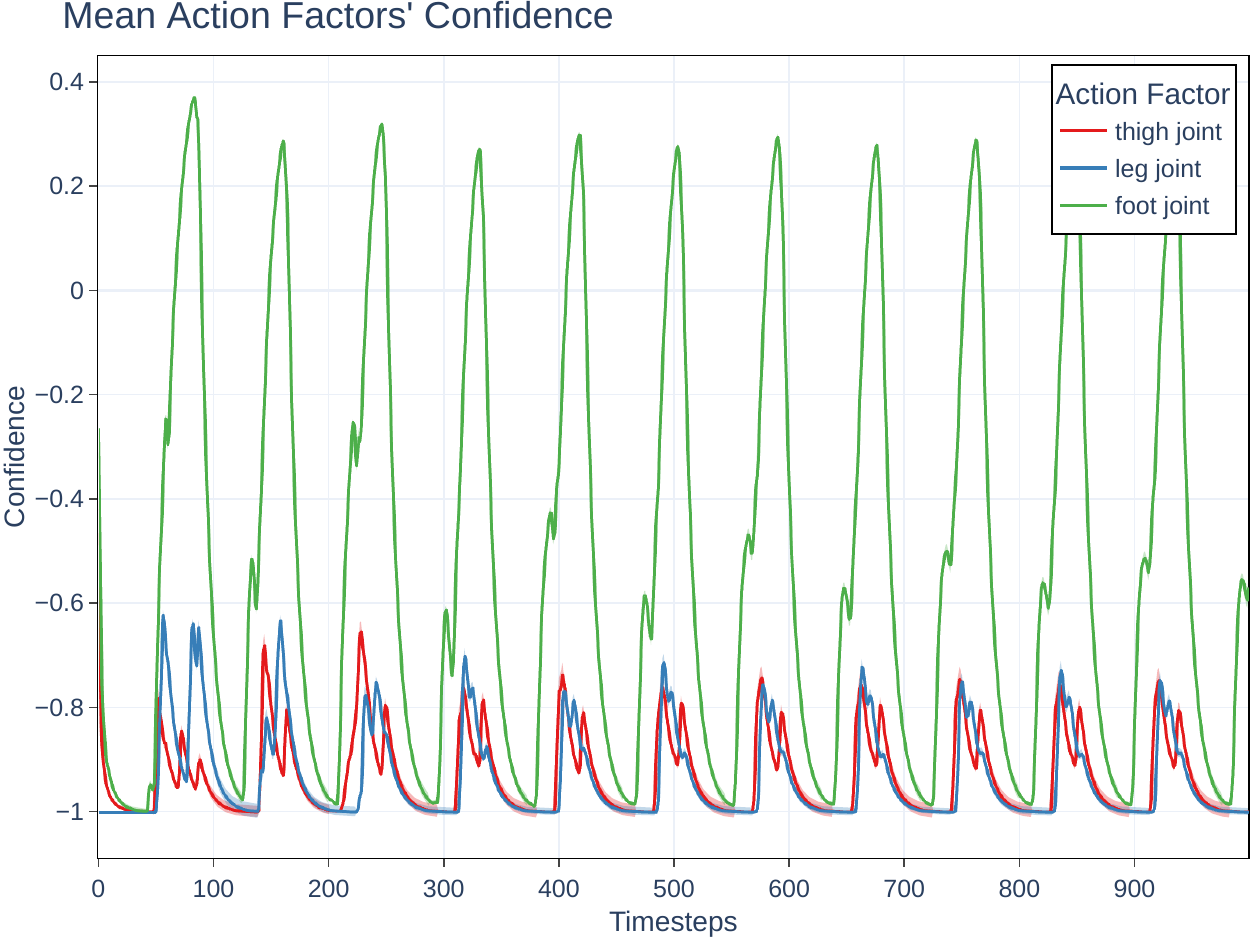}
        \caption{Confidence over time.}%
        \label{Fig:HopperConfTime}
    \end{subfigure}
    \caption{Results of interestingness analysis for the Hopper scenario.}
	\label{Fig:HopperResults}
\end{figure*}

While this analysis allows us to gain insights about the characteristics of each task and provides an overview of each agent's competence, we can use our framework to assess competence at a deeper level. Since Hopper generates interestingness for all dimensions, we use it here as an example to analyze the agent's competence over time and across different actuators. Fig.~\ref{Fig:IntRadars} shows that the agent has on average very low confidence in selecting actions (mean $\mathcal{C}=-0.8\pm0.27$) but Fig.~\ref{Fig:HopperConfBar} shows that confidence varies greatly with the action factor. Here, we see that the agent is more confident about selecting values for the \emph{foot joint} actuator compared to the others. Fig.~\ref{Fig:HopperConfTime} shows another level of detail by depicting how Confidence varies over the length of a trace for each action factor. It reveals a cyclic pattern where the agent is actually quite confident at times in selecting values for the \emph{foot joint} (peaks at $\mathcal{C}\approx0.3$). An empirical analysis of the videos associated with the traces reveals that Confidence increases when the agent is preparing to jump and reaches its lowest values when the agent is in mid-air, which makes sense since at that point controlling the foot joint has little effect. Overall, we can see that depending on the scenario and agent's policy, interestingness can vary over time and across different action factors. Next, we investigate whether interestingness can be used to identify different sub-tasks that might denote distinct behaviors and conditions.

\subsection{Trace Clustering based on Interestingness}%
\label{Subsec:Clustering}

With the goal of using interestingness to identify distinct, meaningful patterns---which would indicate that an agent has achieved some level of competence---we clustered agents' traces using only the interestingness data. Our hypothesis is that each cluster will highlight different behaviors and represent distinct aspects of competence as captured by interestingness. For the purpose of this investigation we used the data generated with the Breakout scenario.
%
Since each interestingness analysis computes a value for each timestep of a trace, the result is a set of sequences of numeric data. Since we noted a high variance in the length of the Breakout traces, to allow clustering the traces we first computed the mean value of each interestingness dimension for each trace. We then computed the Euclidean distances between each pair of traces,%
\footnote{For traces with similar length, alternative methods such as Dynamic Time Warping (DTW) \citep{salvador2007dtw} could be used to align and compute the distances between traces.}
which were then fed to a Hierarchical Clustering \citep{kaufman1990agglomerative} algorithm with a complete linkage criterion.  

\begin{table}[!ht]
    \centering
    \caption{Characteristics of the trace clusters found for the Breakout scenario.}
    \begin{tabular}{l@{\hspace{10pt}} 
        r@{\hspace{10pt}}
        r@{\hspace{2pt}$\pm$\hspace{2pt}}r@{\hspace{10pt}} 
        r@{\hspace{2pt}$\pm$\hspace{2pt}}r@{\hspace{10pt}}
        r@{\hspace{2pt}$\pm$\hspace{2pt}}r@{\hspace{10pt}}}
    \toprule
        \textbf{Cluster} & 
            \multicolumn{1}{c}{\textbf{Size}} & 
            \multicolumn{2}{c}{\textbf{Lives}} & 
            \multicolumn{2}{c}{\textbf{Score}} & 
            \multicolumn{2}{c}{\textbf{Length}} \\ 
    \midrule
        \textbf{0} & $451$ & $3.53$ & $0.75$ & $11.53$ & $13.47$ & $247.92$ & $238.26$ \\ 
        \textbf{1} & $226$ & $3.72$ & $0.45$ & $318.42$ & $28.82$ & $2878.58$ & $1131.26$ \\ 
        \textbf{2} & $70$ & $1.00$ & $0.00$ & $359.00$ & $0.00$ & $740.00$ & $0.00$ \\ 
        \textbf{3} & $190$ & $0.33$ & $0.47$ & $29.28$ & $15.23$ & $331.82$ & $181.09$ \\ 
        \textbf{4} & $63$ & $2.00$ & $0.00$ & $37.00$ & $0.00$ & $166.00$ & $0.00$ \\ 
    \bottomrule
    \end{tabular}
    \label{Table:BreakoutClusters}
\end{table}

\begin{figure}[!tb]
    \centering
    \includegraphics[width=0.7\textwidth]{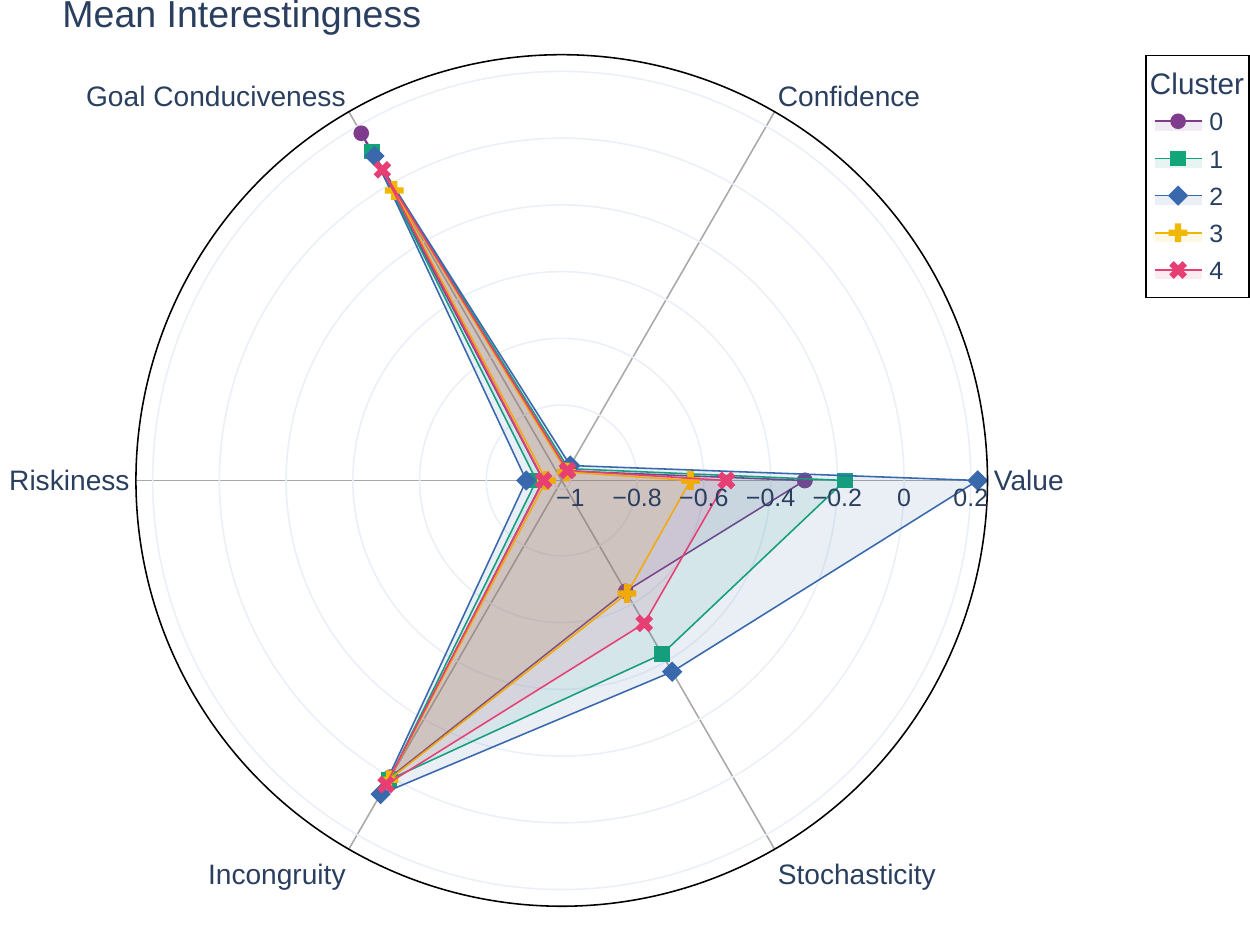}
    \caption{Interestingness profiles for each trace cluster in the Breakout scenario.}
	\label{Fig:BreakoutClustersInt}
\end{figure}

To select the number of clusters, we compute the Silhouette coefficient \citep{rousseeuw1987silhouette}. While the best partition according to this metric resulted in $14$ clusters, for illustrating the results of clustering by interestingness we chose $5$ clusters since it provided the best balance between number of clusters and cluster sizes. Table~\ref{Table:BreakoutClusters} shows various agent performance metrics and characteristics of the traces in each resulting cluster. \textbf{Score} corresponds to the cumulative reward collected by the agent \emph{during} the trace, which might be different from the total game (in Breakout, a trace ends whenever the agent loses a life). As we can see, each cluster captures different competency-controlling conditions and different stages of the Breakout game. This table is complemented by Fig.~\ref{Fig:BreakoutClustersInt} showing the different interestingness profiles resulting for each cluster, where we see that profiles differ mainly in the Value and Stochasticity dimensions. Together with an analysis of the traces' videos in each cluster, this provides an understanding of the distinct challenges faced by the agent. 

In particular, cluster $0$ is characterized by the agent being in an early stage of the Breakout game where many lives are still available and the agent is unable to make much progress (low score, short trace lengths). Cluster $1$ captures situations occurring in early stages of the game similar to cluster $0$ but where the agent is able to make good progress on the task, which translates to the agent achieving a very high score, resulting in longer traces. In Fig.~\ref{Fig:BreakoutClustersInt}, we see that the agent assigns a higher Value on average to cluster $1$ compared to cluster $0$. As for cluster $2$, all traces capture the exact same strategy of destroying the bricks on both sides of the wall, making the ball bounce on the top part of the environment and destroying the wall from above, resulting in the highest average score among clusters. This cluster is also characterized by the highest Value among clusters as well as the highest Stochasticity, presumably because the agent's actions have no effect while the ball is bouncing on the top, so the agent receives different rewards (whenever the ball hits a brick) on seemingly the same situation, resulting in stochasticity as experienced by the agent. In contrast, cluster $3$ occurs at late stages of the game where there are only a few bricks left and the agent eventually fails to destroy all of them before losing its last remaining lives. As seen in Fig.~\ref{Fig:BreakoutClustersInt}, this is also the cluster with the lowest associated mean Value. Finally, cluster $4$ captures situations where the behavior of the agent follows the exact the same pattern in each trace: the agent already has a relatively high game score, there are a few bricks left, and the agent loses a life when the ball is deflected on the right side of the environment near the bottom.

Overall, the scenarios in some clusters are indicative of situations where agent behavior can be improved, while others highlight situations where the agent behaves optimally (achieves its objectives), thereby denoting additional competency-controlling conditions. Furthermore, we note that the clusters were discovered using solely the interestingness analyses and did not rely on any external information such as game scores or remaining lives.

\subsection{Global and Local Analysis of Feature Importance}%
\label{Subsec:FeatureImportance}

So far, we have showed how we can use interestingness to better understand how interestingness correlates with agent behavior patterns and how we can cluster traces based on interestingness to identify different sub-tasks within a scenario, denoting distinct competency-controlling conditions. Feature importance analysis allows us to gain deeper insight into which task elements most affect an agent's competence, how they affect the agent's behavior as measured by interestingness, and where (in which situations) this occurs. 

For this analysis we use the SC2 Combat scenario since it is the one where the elements of the task (different buildings, unit types) could affect interestingness the most. To perform feature importance analysis, we first need a set of high-level, interpretable features whose influence on each dimension of analysis we can predict and explain. Given that the \texttt{pySC2} interface provides semantic but still relatively low-level information, we utilized the high-level SC2 feature extractor described in \citep{sequeira2022camly2}, which provides numeric descriptions for the task elements (types and properties of units, amount and size of forces, etc.) and abstracts the agent's local behavior (movement of groups of units relative to the opponent, the orders assigned to each unit type, etc.).

To identify which features impact each interestingness dimension the most, we used SHapley Additive exPlanations (SHAP) \citep{lundberg2017shap}. SHAP values provide explanations by computing the impact of each feature's value on the prediction of a target variable, given a trained predictive model. 
%
For ease of analysis and SHAP computation, we used XGBoost (eXtreme Gradient Boosting) machines \citep{chen2016xgboost}, which minimize the loss iteratively by adding weak learners (in our case, small decision trees) using a gradient-descent-like procedure. We trained regressors from our SC2 numeric high-level features ($529$ total) to each of the $13$ interestingness dimensions using RMSE loss. We used the $1{,}000$ traces sampled from our SC2 agent, randomly selecting $80\%$ of the timesteps to get $\approx13\times10^5$ training instances. By looking at the mean absolute error of the models tested on the remaining $20\%$ of the data, we observed that most models achieve good prediction accuracy except for Goal Conduciveness and Incongruity,%
\footnote{Because these dimensions rely on information from multiple timesteps, a more robust model, making use of past information, is likely required to provide good predictions.}
so we refrain from using these models for feature importance analysis.

\begin{figure}[!tb]
    \centering
    \includegraphics[width=0.75\textwidth]{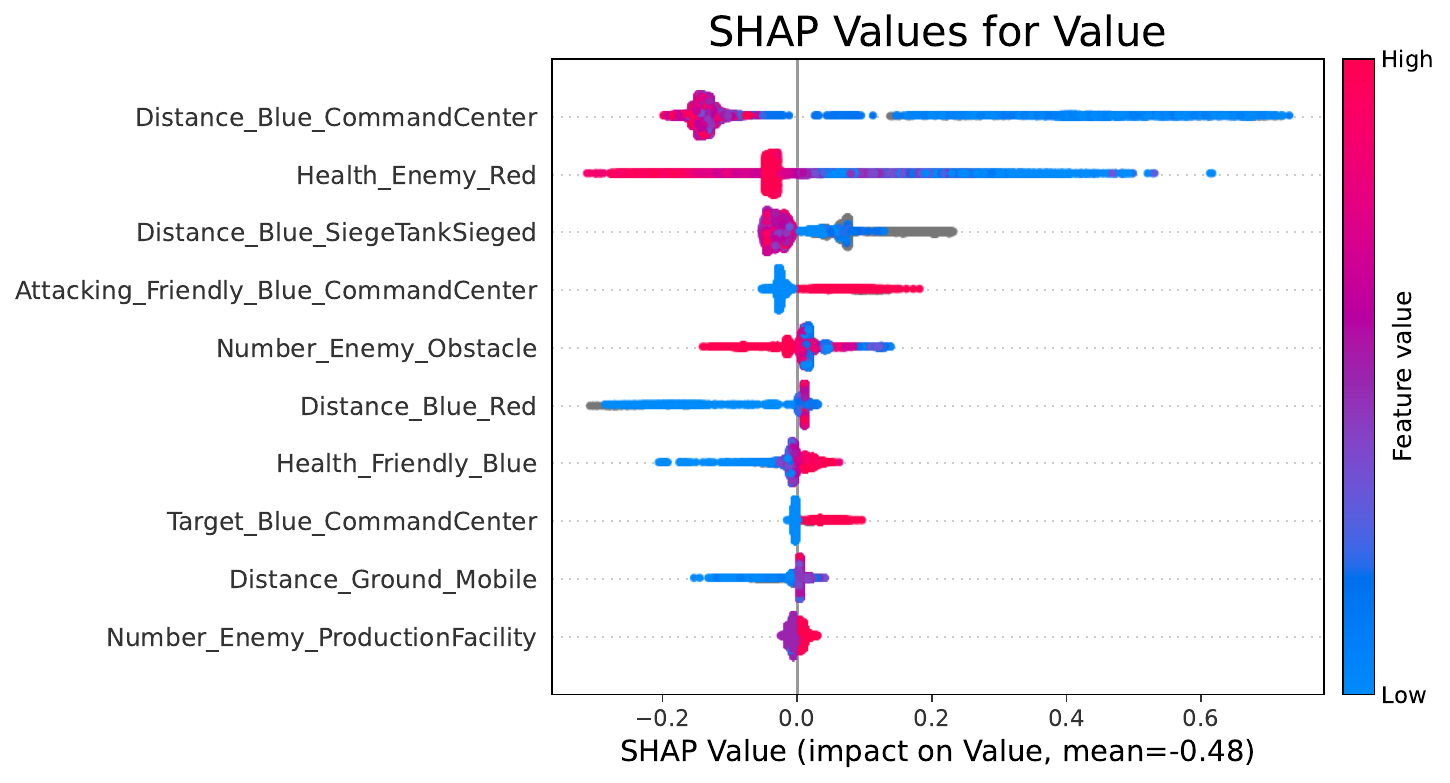}
    \caption{SHAP density plot for Value.}%
    \label{Fig:SHAPValueDensity}
\end{figure}

\subsubsection{Global Interpretation:}

The goal of the global interpretation of interestingness is to understand how distinct aspects of the task influence interestingness in general. We computed the mean SHAP values for the test set ($\approx3\times10^5$ instances), which provides a good estimate of the features' importance for the prediction of each interestingness variable. As an example, Fig.~\ref{Fig:SHAPValueDensity} shows the SHAP density plot for the model of Value for the $10$ most predictive features (y-axis). Colored dots represent datapoints (prediction instances), stacked vertically when they have a similar SHAP value for a feature (x-axis). The color represents the corresponding feature's value, from blue (low feature values) to red (high feature values). Together, they represent how much a feature's value contributes to the prediction of Value, relative to the mean Value. As we can see, the top features show that the distance to CC (and enemy in general), the health of both forces, and whether the agent is attacking the CC have the most impact on the predicted magnitude of Value. We also see that when the agent is attacking the CC, the closer it is, the higher the Value; and that the healthier the agent / the weaker the enemy is, the higher the Value. 

\subsubsection{Local Interpretation:}

\begin{figure*}[!tb]
    \centering
    \begin{subfigure}[b]{0.38\textwidth}
        \includegraphics[width=\textwidth]{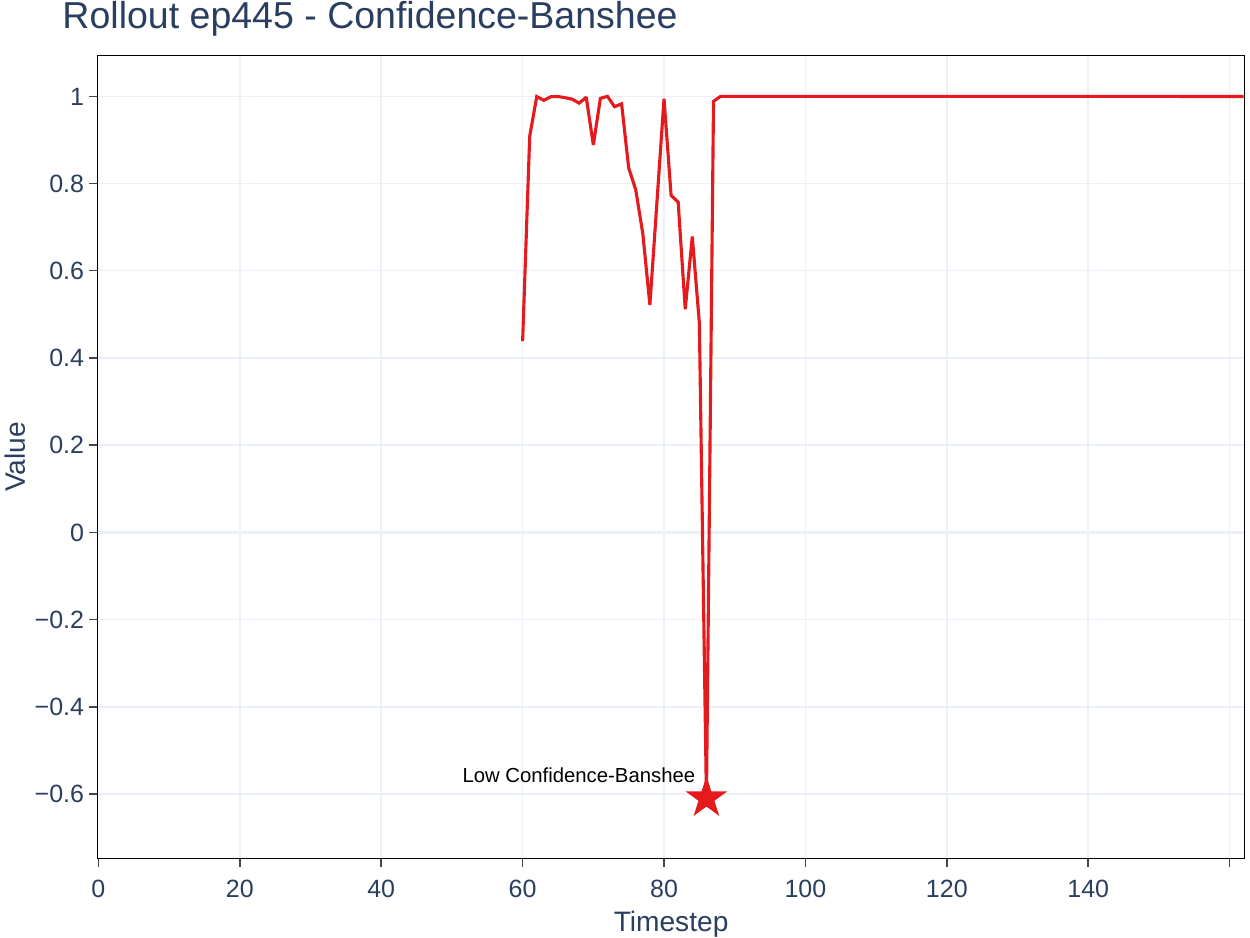}
        \caption{}%
        \label{Fig:LowConfidenceBanshee}
    \end{subfigure}\hspace{5pt}%
    \begin{subfigure}[b]{0.6\textwidth}
        \includegraphics[width=\textwidth]{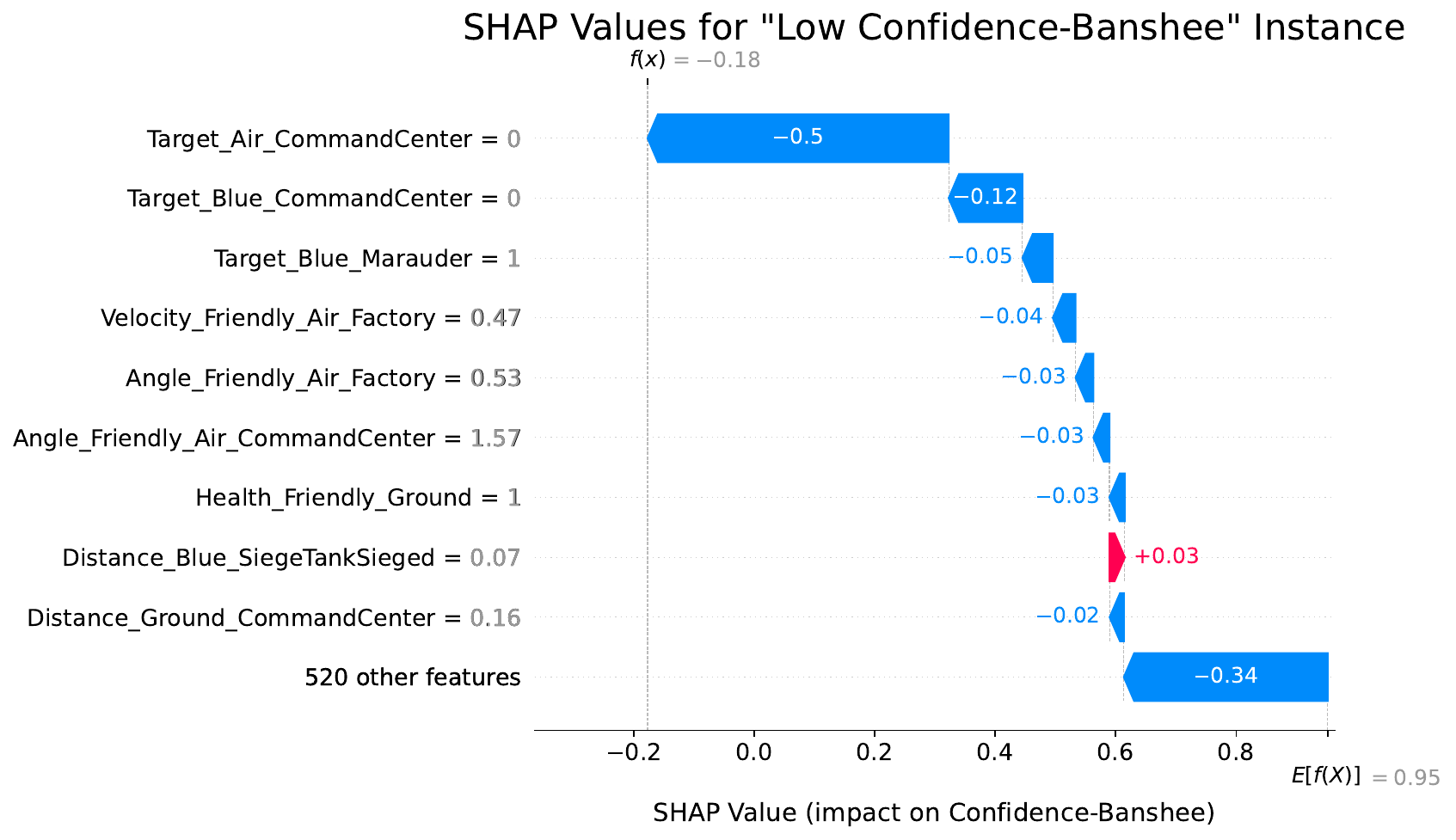}
        \caption{}%
        \label{Fig:SHAPLowConfidenceBanshee}
    \end{subfigure}
    \caption{Local interpretation for a trace with low Confidence in using Banshees: (a) the Confidence in using Banshees throughout the trace; (b) SHAP values for the low Confidence situation.}
	\label{Fig:LocalInterpretation}
\end{figure*}

Global interpretation allows us to identify the task elements influencing an agent's task competency the most, indicating possible 
factors influencing the agent's capabilities and limitations. However, as discussed earlier, a good competency-aware system needs to model the situations one would expect a human operator to be curious about, ones in which a knowledge gap occurs and where further information is needed to make sense of the event \citep{hoffman2018xai}. These provide opportunities for self-explanation, requesting feedback, counterfactual reasoning, etc. 
%
As such, we use local interpretation of interestingness to understand particular ``key'' moments for the agent. Specifically, for each dimension, we identified the timesteps (across all the $1{,}000$ sampled traces) where the interestingness value was significantly different from the average (identified via the interquartile range method). Then, for each situation deemed ``abnormal,'' we computed the SHAP values to identify which task elements might be responsible for the agent's high or low task competency. 

Here we illustrate the usefulness of local interpretation with one example. Fig.~\ref{Fig:LowConfidenceBanshee} plots the agent's Confidence in using Banshees for one trace, where we observe a sudden drop at around timestep $85$. From previous results, we know that the agent is usually highly confident about using Banshees. Watching the episode's replay reveals that this happened when the aerial units stopped targeting the CC and turned away from it (in the direction of the Factory) and the Blue units started targeting the Red Marauders. This information is also reflected in the SHAP values for each feature, as illustrated in the waterfall plot in Fig.~\ref{Fig:SHAPLowConfidenceBanshee}, which indicates how the value of the $10$ most impactful features (y-axis) contributed to the large deviation of Confidence from the mean (x-axis). This explains why the agent suddenly had low confidence in using the Banshees: even though Banshees usually target the CC alone since they cannot be defeated, here they were ``forced'' to help the ground units. Although this particular situation may be rare, it may still lead to undesirable consequences, \eg to the loss of ground units.

By analyzing other examples of (extremely) low Confidence in using Banshees, we noted patterns in the SHAP values of those instances, as well as observed similar scenarios when watching the corresponding replays. In the future, we will automatically identify these patterns \eg by clustering the outliers based on the SHAP values for a given dimension. Additionally, end-users of our system could use the patterns to recommend alternative courses of action or remedy the agent's policy by guiding it during deployment, \eg by suggesting that the agent's ground units not engage the enemy when the agent has aerial units at its disposal. An alternative human intervention could be to retrain the agent \eg by modifying its reward function, such that when Banshees are present, ground units never engage the enemy, or, better yet, move to a safer location.

\section{Conclusions}%
\label{Sec:Conclusions}

In this paper, we presented a novel framework for characterizing the competence of deep RL agents based on interestingness analysis. Competency assessment is done by analyzing an agent's behavior along seven distinct interestingness dimensions: Value, Confidence, Goal Conduciveness, Incongruity, Riskiness, Stochasticity and Familiarity. Each dimension captures competence along a distinct aspect of the agent's interaction with the environment, making use of internal information by probing the models that are optimized by the algorithm during RL training. Our implementation supports a wide range of deep RL algorithms and is natively compatible with popular RL toolkits.

We conducted a computational study where we trained agents for different game scenarios using different RL algorithms. We applied our framework to each trained agent by extracting interestingness data from traces produced by each agent policy. We then used different methods to analyze the agents' competence based on interestingness. 
Our results show that different RL agents trained on distinct scenarios leads to different interestingness profiles and that analyzing temporal patterns of interestingness across traces reveals how an agent's behavior correlates with interestingness. In addition, we demonstrated how clustering traces based only on interestingness enables the discovery of distinct behaviors, applied under different competency-controlling conditions. Global interpretation of feature importance shows how we can identify which (and how) task factors affect an agent's competence along each dimension. Finally, local interpretation helps determine the contribution of each task element of a particular scene---identified as having an ``abnormal'' interestingness value---which allows gaining insights into an agent's limitations, and helps identify measures that might need to be taken to improve its performance.

Our current framework is directed at agent designers who are RL experts by providing a more holistic view of an RL agent's competence, without which it is harder to understand an agent's capabilities and limitations, identify potential barriers for optimal performance, or realize what interventions are needed to help the agent---and its human operators and teammates---achieve their goals. Currently, in addition to exploring ways to make better use of the interpretation mechanisms, we are developing user interfaces to facilitate the use of our competency awareness tool. Namely, since all of our interestingness analyses can be computed online based solely on interaction data collected up to the current timestep, we are exploring ways in which our tools can be used to guide (non-technical) operators during agent deployment, alerting them in situations in which operator-specified conditions or goals would be violated, and requesting their input in order to improve the agent's performance.

\section*{Acknowledgements} This material is based upon work supported by the Defense Advanced Research Projects Agency (DARPA) under Contract No. HR001119C0112. Any opinions, findings and conclusions or recommendations expressed in this material are those of the author(s) and do not necessarily reflect the views of the DARPA.

\bibliographystyle{ACM-Reference-Format}
\bibliography{23-xai-ixdrl}

\end{document}